# Assessment of Anterior Cruciate Ligament Injury Risk Based on Human Key Points Detection Algorithm

Ziyu Gong[1], Xiong Zhao[2], Chen Yang[3,4]

[1]*College of Computer Science and Technology, National University of Defense Technology, Changsha, 410073, China*

[2]*School of Human Kinetics, Faculty of Health Sciences, University of Ottawa, 200 Lees Avenue, Ottawa, Ontario K1N 6N5, Canada*

[3]*Department of Physical Medicine and Rehabilitation, Northwestern University, 710 N Lake Shore Dr, Chicago, IL, 60611, USA*

[4]*Max Nader Lab for Rehabilitation Technologies and Outcomes Research, Shirley Ryan AbilityLab, 355 E Erie St., Chicago, IL, 60611, USA*

## Abstract

This paper aims to detect the potential injury risk of the anterior cruciate ligament (ACL) by proposing an ACL potential injury risk assessment algorithm based on key points of the human body detected using computer vision technology. To obtain the key points data of the human body in each frame, OpenPose, an open-source computer vision algorithm, was employed. The obtained data underwent preprocessing and were then fed into an ACL potential injury feature extraction model based on the Landing Error Evaluation System (LESS). This model extracted several important parameters, including the knee flexion angle, the trunk flexion on the sagittal plane, trunk flexion angle on the frontal plane, the ankle-knee horizontal distance, and the ankle-shoulder horizontal distance. Each of these features was assigned a threshold interval, and a segmented evaluation function was utilized to score them accordingly. To calculate the final score of the participant, the score values were input into a weighted scoring model designed based on the Analytic Hierarchy Process (AHP). The AHP-based model takes into account the relative importance of each feature in the overall assessment. The results demonstrate that the proposed algorithm effectively detects the potential risk of ACL injury. The proposed algorithm demonstrates its effectiveness in detecting ACL injury risk, offering valuable insights for injury prevention and intervention strategies in sports and related fields. Code is available at: https://github.com/ZiyuGong-proj/Assessment-of-ACL-Injury-Risk-Based-on-Openpose

## Introduction

Improper movement patterns can highly increase the risk of injury and are detrimental to the quality of exercise. For athletes, a fixed movement pattern often results in wear and tear on certain soft tissues or joints and increases the risk of injuries. Therefore, screening out potential injuries is of great significance to athletes and sports enthusiasts. Movement pattern assessment has become a common method to identify potential sports injury risks [1]. LESS and functional movement screening system (Functional Movement Screen, FMS) are frequently used to identify human movement patterns. However, the LESS and FMS usually rely on the visual evaluation of the clinician or trainer through video playback. The process can be subjective and time-consuming [2]. Mauntel et al. [3] used the 3D depth camera Kinect to achieve automatic scoring of LESS, but the automation of LESS using 2D camera is still not available. The increasing popularity of utilizing 2D cameras in motion analysis is attributed to the advancements in computer vision, which have led to improved accuracy.

This study implements the automatic scoring function of the LESS by utilizing human body key points data generated by OpenPose from a 2D camera. [4]. To begin with, we introduce a method for extracting features related to the potential injury risk of the ACL based on the key points of the human body. Subsequently, utilizing the ACL potential injury risk scoring function outlined in this paper, the extracted features are assigned scores. The process involves initially utilizing OpenPose to capture the key points data of the human body in each image frame. The data is then preprocessed and input into the feature extraction model to obtain the corresponding feature values, including the knee flexion and trunk flexion angles in the sagittal plane, the lateral trunk angle in the frontal plane, as well as the ankle-knee horizontal distance and the ankle-shoulder horizontal distance.

Using the defined scoring function, the extracted feature values undergo scoring operations based on distinct threshold intervals assigned to each feature. Building upon the five characteristic values obtained in the previous step, a weighted scoring model, developed through the analytic hierarchy process (AHP), is employed to calculate the participant's final score. The individual score and final score are then combined to comprehensively evaluate the participant's risk.

Finally, the analysis of OpenPose's recognition accuracy and the experimental results of the ACL potential injury risk assessment method were conducted. The author acknowledges the significant innovation and practical value of this research in the field of movement evaluation.

## Related work

In 2009, Padua et al. [2] first proposed the LESS to identify subjects with a higher risk of ACL injury. The test process was recorded using cameras positioned in the sagittal and frontal planes to capture the entirety of the evaluation. The subjects' scoring was performed manually by observing the recorded videos. The study revealed significant differences in lower limb kinetics between subjects with low LESS scores (indicating excellent landing techniques) and those with high LESS scores (indicating poor landing technique). Therefore, it is concluded that LESS is a reliable kinematic evaluation method that can be used to identify the potential risk of injury during the landing process.

Clark et al., 2012 [5] conducted a study to assess the effectiveness of Kinect for posture evaluation. The research involved administering three balance tests to a group of 20 healthy subjects. The study's findings indicated that Kinect can be successfully utilized for human posture evaluation. This research holds significance in demonstrating the potential value of integrating Kinect depth cameras into human kinematics analysis. In a study conducted by Schmitz et al., 2015 [6], a comparison was made between a marker-based motion capture system and a markerless motion capture system utilizing the Kinect depth camera. The investigation specifically focused on analyzing the peaks of flexion angles of lower limb joints during squatting exercises performed by 15 subjects. The results demonstrated a high correlation between the two systems, indicating the feasibility and reliability of the markerless motion capture system based on the Kinect depth camera for capturing lower limb joint angles. In a similar fashion, Perrott et al., 2017 [7] conducted a comparison between the marker-based and the markerless motion capture system, examining 13 clinically relevant joint angles during the squatting motion. Their findings indicated that in 9 out of the 13 joint angles, there was no significant difference between the two systems. This suggests that the markerless motion capture system, based on the Kinect depth camera, can provide comparable results to the marker-based system for assessing joint angles during squatting exercises.

In 2017, Mauntel et al. [3] introduced automated enhancements to the LESS test method by leveraging the Kinect depth camera, effectively addressing the research gap in automated injury assessment within the field of movement evaluation. Furthermore, Mentiplay et al. [8], in their 2018 study, utilized the Kinect depth camera to measure the kinematics of the hips and knees in both frontal and sagittal planes during single-leg squat (SLS) and drop vertical jump (DVJ) exercises. The findings indicated that the Kinect depth camera holds potential in screening for ACL injury risk. Additionally, in the 2019 study by Clark et al., 2019 [9], they explored the quest for an optimal movement assessment method and alternative approaches to Kinect. Their work emphasized the future development of a movement assessment system based on OpenPose, which holds implications for the present research.

This article focuses on the automatic evaluation of the LESS using a 2D camera and OpenPose. To extract key points, the bottom-up multi-person pose estimation approach initially requires identifying the position of an individual within the image using a bounding box. Subsequently, the pose of the person within the bounding box is estimated. In the initial stages of this research field, Ladicky et al. [10] employed HOG features for segmenting and locating different parts of the human body. Gkiocari et al. [11] proposed a method based on K-poselets for detecting human body key points. Pischulin et al. [12] utilized deep features to predict joint positions and applied the Deepcut clustering method for joint point classification. However, this approach suffered from increased algorithmic time complexity, resulting in slower processing speeds. In a subsequent study, Insafutdinov et al. [13] introduced the ResNet framework for human body part detection and incorporated incremental optimization strategies to enhance the computational efficiency of the Deepcut algorithm. Furthermore, Cao et al. [4] developed the OpenPose multi-person pose estimation open-source library. This library utilizes a network with two stacked and cross-linked convolution branches, where one branch predicts key points, and the other branch infers limb relationships. The real-time generation of key points data by OpenPose renders it highly practical and valuable for automating the LESS.

## Methods

### *ACL Injury feature extraction model*

Figure 1 illustrates the flow chart of the proposed potential injury assessment system based on human body key point data in this paper. The process begins by obtaining and processing Body_25 human body key point data in each frame using OpenPose. Next, the system calculates the five feature values of the human body. Based on their respective threshold scoring functions, the individual test values are assigned scores. Specifically, a score of 9 denotes excellent performance, 5 represents good performance, and 1 indicates poor performance. Subsequently, the calculated five features are fed into the Analytic Hierarchy Process (AHP) weight evaluation model, resulting in the final evaluation score.

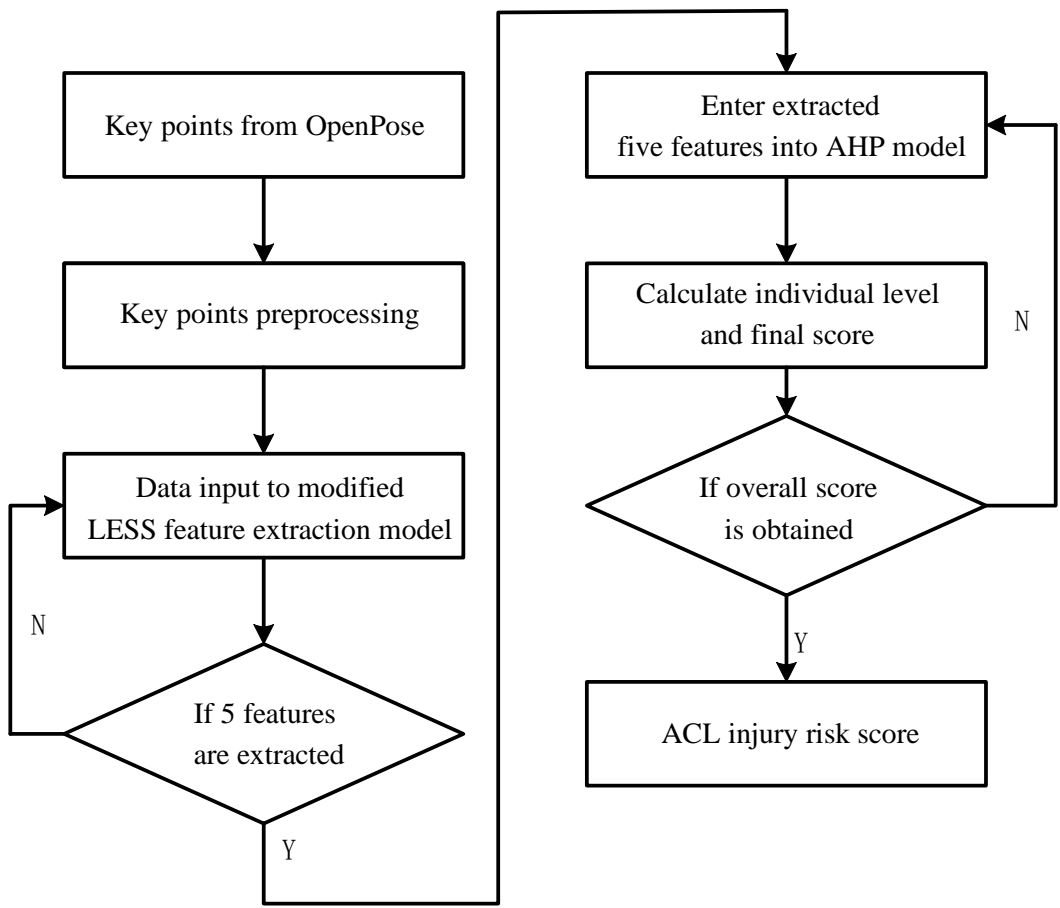

Figure 1: Flow chart of potential injury assessment system based on human body key points data.

Following the evaluation and consolidation of the 17 scoring rules in the LESS, it was determined that four out of the 17 individual scoring indicators pertained to assessing the subject's feet. Taking into account the recognition accuracy of OpenPose [14], five evaluation indicators were selected to specifically identify the potential risk of ACL injury. In conjunction with the study conducted by Sinsurin et al. [15], it was observed that an increase in the flexion angles of the hip and knee in the sagittal plane can effectively mitigate the risk of injury. Ameer et al. [16] revealed that increased peak knee flexion during single-leg landings may serve as a protective mechanism against ACL injury. Specifically, when the knee flexion angle is less than 30°, the strain on the ACL is significantly higher compared to when the knee flexion angle is larger. On the other hand, when the knee flexion angle exceeds 60°, the load on the ACL is considerably reduced. Moreover, increasing the knee flexion angle after landing can effectively reduce the impact force [17], consequently reducing the risk of ACL injury.

This paper uses the platform landing support (Drop Landing, DL) test to extract the aforementioned five features, and employed the modified LESS scoring criteria to evaluate the ACL potential injury risk as shown in Table 1.

Table 1: ACL potential injury risk assessment index.

| Scoring item | Criterion | Score |
|---|---|---|
| A1: Sagittal plane: Peak flexion angle of right thigh and shank | A peak value between 0 and 30 degrees represents poor; A peak value between 30 degrees and 60 degrees represents good; A peak value greater than 60 degrees represents excellent | 1=poor 5=good 9=excellent |
| A2: Sagittal plane: Peak flexion angle of right thigh and trunk | A peak value between 0 and 30 degrees represents poor; A peak value between 30 degrees and 60 degrees represents good; A peak value greater than 60 degrees represents excellent | 1=poor 5=good 9=excellent |
| A3: Frontal plane: the average of peak values of the flexion angles between the two thighs and the trunk | A peak value greater than 60 degrees means poor; A peak value between 60 degrees and 30 degrees means good; A peak value between 0 degrees and 30 degrees means excellent | 1=poor 5=good 9=excellent |
| D1: Frontal plane: the absolute value of the maximum difference between the horizontal distance between the two knee joints and the horizontal distance between the two ankles | A value greater than 50 means poor; A value between 30-50 means good; A value less than 30 means excellent; | 1=poor 5=good 9=excellent |
| D2: Frontal plane: the absolute value of the maximum difference between the width of the shoulders and the width of the feet | A value greater than 50 means poor; A value between 30-50 means good; A value less than 30 means excellent; | 1=poor 5=good 9=excellent |

*Sagittal plane feature extraction model*

First, this article defines the time series frame $T$ and the sagittal feature set $P$, as shown in equation 1 and equation 2:

$$\text{Time series frame：} T = \{t_1, t_2, t_3 \cdots t_n\}, n > 0 \quad (1)$$

$$\text{Sagittal features：} P = \{p_1, p_2\} \quad (2)$$

Based on the LESS sagittal plane scoring items, this paper uses $p_1$ to represent the peak value of the complementary cosine of the knee flexion angle during the initial landing phase. The thigh vector is represented by $\alpha$ and the shank vector is represented by $\beta$, as shown in

Figure 2. And $p_2$ is used to represent the peak value of the cosine value of the supplementary angle of the torso and thigh flexion angle, and the torso vector is defined $\gamma$.

The model is constructed using the key points data generated by OpenPose, as well as the variable $p_1$ and $p_2$ during the DL test. The knee angle feature and the feature representing the angle between the trunk and the thigh are as shown in equations 3 and 4:

$$p_1 = \max\{\frac{\alpha^*\beta}{|\alpha|*|\beta|}\} \cdot \text{s.t.} \cdot \{\alpha,\beta | \alpha = (x_9 - x_{10}, y_9 - y_{10}), \beta = (x_{11} - x_{10}, y_{11} - y_{10})\} \quad (3)$$

$$p_2 = \max\{\frac{\alpha^*\gamma}{|\alpha|*|\gamma|}\} \cdot \text{s.t.} \cdot \{\alpha,\gamma | \alpha = (x_9 - x_{10}, y_9 - y_{10}), \gamma = (x_8 - x_1, y_8 - y_1)\} \quad (4)$$

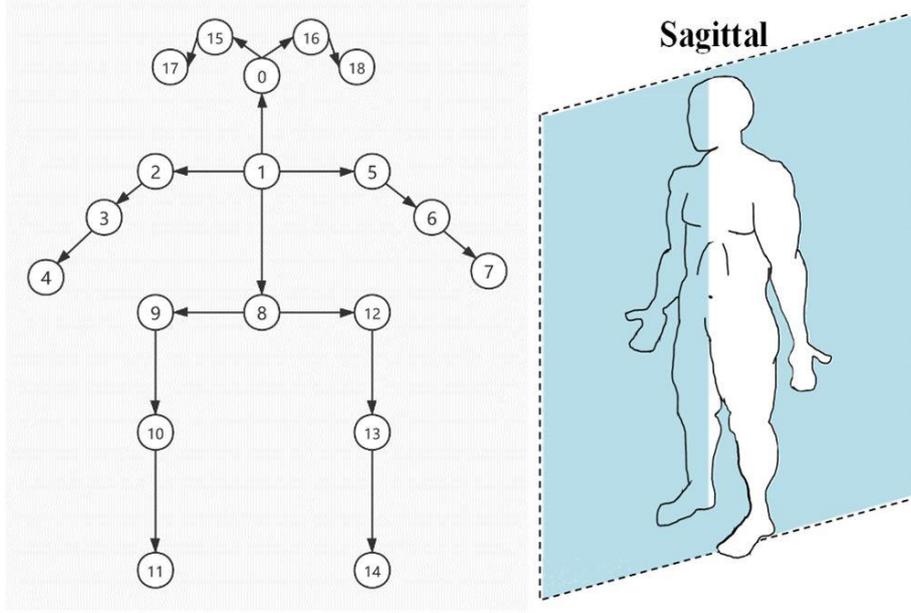

Figure 2: Illustration of the model and the sagittal plane.

The segment scoring functions, denoted as $f(t_n, p_1)$ and $f(t_n, p_2)$, are defined individually. These functions signify that when the time frame is $t_n (n > 0)$, the value of $p_1$ (or $p_2$) is evaluated, and the resulting score value is classified based on the segment function and corresponding threshold on the sagittal plane. Specifically, a score of 9 represents excellent performance, 5 denotes good performance, and 1 indicates poor performance. This is illustrated by equation 5 and equation 6:

$$f(t_n, p_1) = \begin{cases} 9, & -\frac{1}{2} < p_1 \leq 1 \\ 5, & -\frac{\sqrt{3}}{2} < p_1 \leq -\frac{1}{2} \\ 1, & -1 \leq p_1 \leq -\frac{\sqrt{3}}{2} \end{cases} \quad (5)$$

$$f(t_n, p_2) = \begin{cases} 9, & -\frac{1}{2} < p_2 \leq 1 \\ 5, & -\frac{\sqrt{3}}{2} < p_2 \leq -\frac{1}{2} \\ 1, & -1 \leq p_2 \leq -\frac{\sqrt{3}}{2} \end{cases} \quad (6)$$

*Frontal plane feature extraction model*

Similarly, this article first defines the time series frame $T$ and the frontal plane feature set $S$, as shown in equations 7 and 8:

$$\text{Time series frame: } T = \{t_1, t_2, t_3 \cdots t_n\}, n > 0 \quad (7)$$

$$\text{Frontal feature: } S = \{s_1, s_2, s_3, s_4\} \quad (8)$$

Equations 9 to 15 describe the calculations used to determine specific measurements on the frontal plane. In these equations, $s_1$ represents the Euclidean distance (i.e., horizontal width) between the two ankles on the frontal plane, $s_2$ means the Euclidean distance in horizontal between the two knee joints on the frontal plane, $s_3$ stands for the shoulder width on the frontal plane, and $s_4$ represents the peak value of the average flexion angle of the two thighs and the trunk on the frontal plane. These equations are used to model the human body key points generated in real time by OpenPose in the frontal plane. For a visual representation of the human body key points in the frontal plane, please refer to Figure 3.

$$s_1 = \sqrt{(x_{14} - x_{11})^2 + (y_{14} - y_{11})^2} \quad (9)$$

$$s_2 = \sqrt{(x_{13} - x_{10})^2 + (y_{13} - y_{10})^2} \quad (10)$$

$$s_3 = \sqrt{(x_5 - x_2)^2 + (y_5 - y_2)^2} \quad (11)$$

$$\text{Trunk vector: } d = (x_8 - x_1, y_8 - y_1) \quad (12)$$

$$\text{Left thigh vector: } e = (x_9 - x_{10}, y_9 - y_{10}) \quad (13)$$

$$\text{Right thigh vector: } f = (x_{12} - x_{13}, y_{12} - y_{13}) \quad (14)$$

$$s_4 = \max\left\{\frac{1}{2}\left(\frac{d*e}{|d|*|e|} + \frac{d*f}{|d|*|f|}\right)\right\} \quad (15)$$

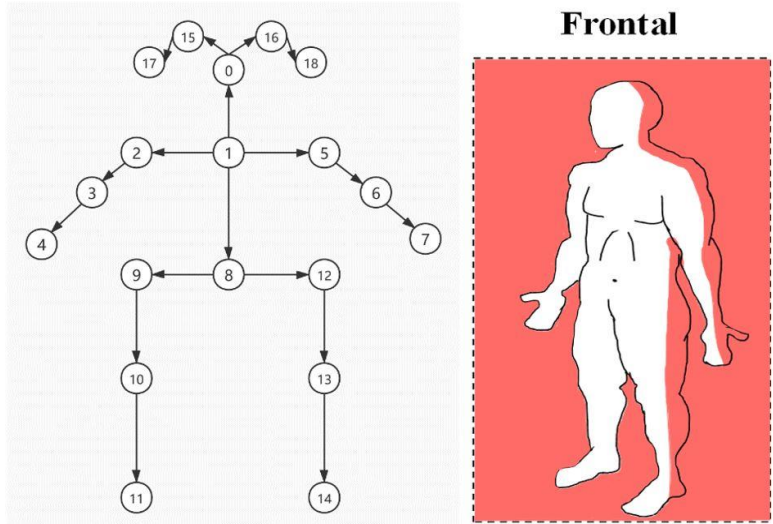

Figure 3: Illustration of the model and the frontal plane.

By combining the frontal features mentioned earlier, this paper establishes segmented evaluation functions $f(t_n, s_1, s_2)$ and $f(t_n, s_1, s_3)$. These functions represent the score values obtained for the features in the $t_n$ frame based on the absolute difference between $|s_1 - s_2|$, as well as $|s_1 - s_3|$, respectively. For the angle-related features, the function $f(t_n, s_4)$ is defined to represent the score value obtained by the feature $s_4$ in the $t_n$ frame. The following equations are defined to construct the model using key point data generated by OpenPose. Similarly, a score of 9 represents excellent performance, 5 denotes good, and 1 indicates poor, as shown in equation 16 to equation 18:

$$f(t_n, s_1, s_2) = \begin{cases} 9, & \max\{|s_1 - s_2|\} < 30 \\ 5, & 30 \leq \max\{|s_1 - s_2|\} < 50 \\ 1, & 50 \leq \max\{|s_1 - s_2|\} \end{cases} \quad (16)$$

$$f(t_n, s_1, s_3) = \begin{cases} 9, & \max\{|s_1 - s_3|\} < 30 \\ 5, & 30 \leq \max\{|s_1 - s_3|\} < 50 \\ 1, & 50 \leq \max\{|s_1 - s_3|\} \end{cases} \quad (17)$$

$$f(t_n, s_4) = \begin{cases} 9, & -1 \leq s_4 \leq -\frac{\sqrt{3}}{2} \\ 5, & -\frac{\sqrt{3}}{2} < s_4 \leq -\frac{1}{2} \\ 1, & -\frac{1}{2} < s_4 \leq 1 \end{cases} \quad (18)$$

***Comprehensive evaluation model***
*Construction of the scoring system*

The Analytic Hierarchy Process (AHP) is a multi-dimensional decision-making method that incorporates both qualitative and quantitative analysis. It allows for judgments to be made based on expert opinions and the experiences of decision makers. The AHP is particularly suitable for conducting multi-index evaluations of complex systems. In the context of evaluating LESS scoring results, it is crucial to establish an objective and rational evaluation system. By utilizing the AHP, subjective judgments and preferences can be quantified and combined with objective measurements to create a comprehensive evaluation framework.

To ensure the consideration of both independence and comprehensiveness of indicators, the AHP is employed in this study to construct a three-level structure comprising the target level, criterion level, and index level when establishing the evaluation index system. A weighted scoring model is established based on the AHP to provide a more visually intuitive representation of the test results. This model incorporates the assigned weights for each criterion and index, allowing for a comprehensive evaluation of the subjects. The results of this weighted scoring model are presented in Figure 4.

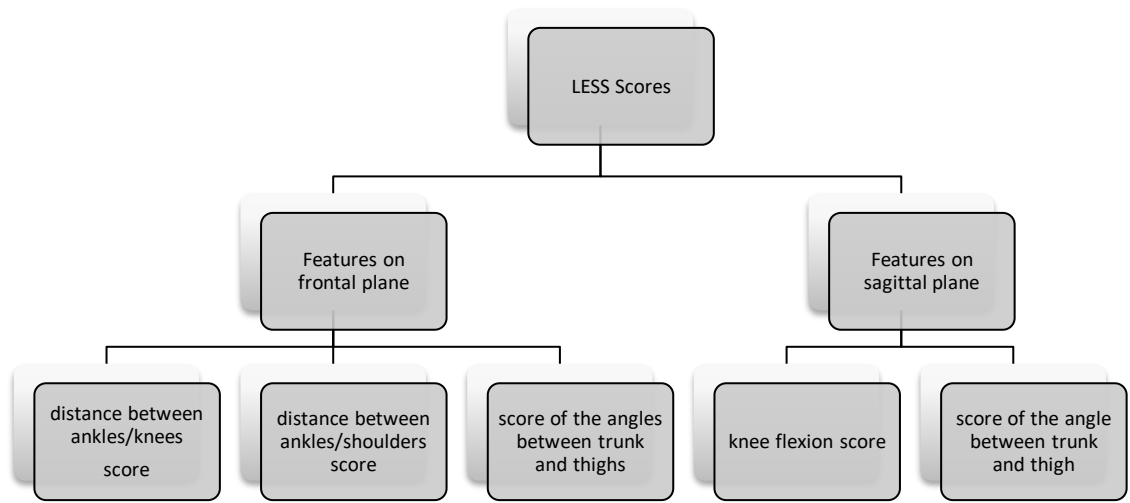

Figure 4: LESS structure chart of scoring results.

*Determination of the weight coefficients of criterion level and index level*

  At the criterion level, the number of indicators is relatively small, making it suitable for a preliminary hierarchical judgment of plan quality. This level allows for intuitive and easily achievable pairwise comparisons. To determine the weights of the indicators at this level, the expert comparison method is utilized. To quantify the comparative indicators, a nine-level scale method is introduced in the evaluation process. This method allows experts to make pairwise comparisons based on the relative importance or preference of the indicators. The nine-level scale provides a structured framework for assigning values to the indicators, facilitating a more systematic and consistent assessment process.

  By employing this method, the evaluation process becomes more rigorous and transparent, enabling experts to effectively compare and prioritize the indicators at the criterion level. This approach ensures a more accurate determination of the weights assigned to each indicator, contributing to the overall evaluation of the plan's quality.

Table 2: AHP evaluation index scale.

| Relative importance scale | Implication |
|---|---|
| 1 | The two elements are equally important |
| 3 | $i$ is slightly important than $j$ |
| 5 | $i$ is obviously important than $j$ |
| 7 | $i$ is much important than $j$ |

| 9 | $i$ is extremely important than $j$ |
|---|---|
| 2,4,6,8 | The median of the above values |
| Reciprocals of above | If activity $i$ has one of the above non-zero numbers signed to it when compared with activity $j$, then $j$ has the reciprocal value when compared with $i$ |

Index $A$ is the parent index of index $B(B_1, B_2 \cdots B_n)$. After comparing and judging by experts and transforming according to the scale of Table 1, the judgment matrix $B$ can be established, and the index weight can be solved by the sum-product method, and finally the consistency test will be carried out as follows:

The judgment matrix $A$ is established by using the relative importance of the indicators, namely:

$$A = (a_y)_{nn} = (w_i/w_j)_{nn} \tag{19}$$

Calculate the square root of $M_i$:

$$M_i = \prod_{j=1}^{n} a_y \ (i,j = 1,2,3,\cdots,n) \tag{20}$$

Normalize the weighted vector $\overline{W}_i$:

$$\overline{W}_i = \frac{\overline{W}_i}{\sum_{j=1}^{n} \overline{W}_j}, (i,j = 1,2,3,\cdots,n) \tag{21}$$

Calculate the maximum eigenvalue $\lambda_{\max}$:

$$\lambda_{\max} = \sum_{i=1}^{n} \left[\frac{(AW)_i}{nW_i}\right] \tag{22}$$

Finally, the consistency check:

$$CI = \frac{\lambda_{\max} - n}{n-1} \tag{23}$$

Consistency check is important to ensure reliable evaluations. It examines the negative mean of remaining characteristic roots and uses the consistency index ($CI$) to assess consistency. The random consistency ratio ($CR$) is compared to a threshold (as in equation 24) to determine if the evaluations pass the consistency test. If not, adjustments are needed to enhance consistency and reliability.

$$\begin{aligned} CR &= CI/RI \\ \text{s.t. } CR &= CI/RI < 0.1 \end{aligned} \tag{24}$$

The values of $RI$ are obtained from Table 3.

Table 3: Random consistency index value.

| $n$ | 1 | 2 | 3 | 4 | 5 | 6 | 7 | 8 | 9 |
|---|---|---|---|---|---|---|---|---|---|
| $RI$ | 0.00 | 0.00 | 0.58 | 0.90 | 1.12 | 1.24 | 1.32 | 1.41 | 1.45 |

*LESS overall score*

Using the weight set $C_i$ of the index layer and the index value matrix $R_i$, the evaluation matrix $B_i$ of the criterion layer can be obtained:

$$\mathbf{B}_i = C_i \times \mathbf{R}_i \tag{25}$$

Finally, the overall evaluation result $P$ can be obtained by multiplying with the weight coefficient of each index of the criterion layer:

$$P = \mathbf{B}_i \times W_i \tag{26}$$

To visually see the evaluation results, the $P$ value corresponds to the evaluation set in Table 4.

Table 4: Index score table.

| Evaluation grade | Evaluation level | Evaluation value |
|---|---|---|
| 1 | excellent | 9 |
| 2 | good | 5 |
| 3 | poor | 1 |

**Verification of LESS scoring results**

*Establishment of index weights at the criterion level*

In this paper, the number of indicators at the criterion level is small, and the method of pairwise comparison is also easy to implement. Therefore, a number of experts are asked to make judgments to determine the judgment matrix at this level, and the judgment matrix $B$ is:

$$\mathbf{B} = \begin{bmatrix} 1 & 3 \\ \frac{1}{3} & 1 \end{bmatrix} \tag{27}$$

The eigenvector of the judgment matrix $B$ is $W = [0.25, 0.75]^T$, the maximum eigenvalue $\lambda_{\max} = 2$, $CI = 0$, and $CR = 0$, which meets the consistency requirements. Therefore, the index weight coefficient of the criterion layer is: (0.25, 0.75).

*Establishing the index weight of the index layer*

Index $B$ is the parent index of index $C$. After experts compare and judge and transform according to the scale, a judgment matrix $R$ can be established, as shown in equation 28:

$$R = \begin{bmatrix} 1 & 2 & 3 & 5 & 5 \\ \frac{1}{2} & 1 & 2 & 3 & 4 \\ \frac{1}{3} & \frac{1}{2} & 1 & 3 & 2 \\ \frac{1}{5} & \frac{1}{3} & \frac{1}{3} & 1 & 2 \\ \frac{1}{5} & \frac{1}{4} & \frac{1}{2} & \frac{1}{2} & 1 \end{bmatrix} \quad (28)$$

The calculated eigenvector of the judgment matrix $W = [0.4267, 0.2574, 0.1602, 0.0886, 0.0671]^T$, the maximum eigenvalue $\lambda_{max} = 5.126$, $CI$ is 0.031, and $CR$ is 0.028<1.12, which meets the consistency requirements. Therefore, the index weight coefficient of the index layer is: (0.4267, 0.2574, 0.1602, 0.0886, 0.0671).

**Participants and protocol**

30 young male participants (10 varsity basketball players, 10 active participants and 10 less active participants) were recruited to test whether there are significant differences in the feature values extracted by the LESS-based ACL potential injury risk feature extraction model between different populations.

The participants were categorized into three groups, and the platform landing support test method was used to extract the five characteristic values from each group. Subsequently, the comprehensive scores for the five individual indicators were calculated using the AHP weighted model, as described in equations (25) and (26). The scoring results are presented in detail in Table 5.

Table 5: Test results of the three groups of subjects.

| Number | X1 | X2 | X3 | X4 | X5 | Total Score |
|---|---|---|---|---|---|---|
| 1 | 9 | 9 | 9 | 9 | 1 | 8.4398 |
| 2 | 9 | 9 | 9 | 1 | 5 | 8.0202 |
| 3 | 9 | 9 | 5 | 1 | 5 | 7.3794 |
| 4 | 9 | 9 | 9 | 9 | 5 | 8.7082 |
| 5 | 9 | 9 | 5 | 9 | 5 | 8.0674 |
| 6 | 9 | 1 | 5 | 9 | 5 | 6.0081 |
| 7 | 9 | 9 | 5 | 1 | 9 | 7.6478 |
| 8 | 5 | 9 | 5 | 9 | 9 | 6.6290 |
| 9 | 5 | 9 | 9 | 9 | 9 | 7.2698 |
| 10 | 9 | 9 | 9 | 9 | 9 | 8.9766 |
| 11 | 9 | 9 | 1 | 1 | 1 | 6.4702 |
| 12 | 5 | 5 | 1 | 1 | 1 | 3.7338 |
| 13 | 9 | 9 | 9 | 9 | 5 | 8.7082 |
| 14 | 5 | 5 | 5 | 1 | 5 | 4.6430 |
| 15 | 5 | 5 | 9 | 9 | 1 | 5.7034 |
| 16 | 1 | 1 | 1 | 1 | 5 | 1.2658 |
| 17 | 5 | 1 | 9 | 5 | 9 | 4.8666 |
| 18 | 1 | 5 | 9 | 9 | 5 | 4.2650 |
| 19 | 5 | 5 | 9 | 5 | 5 | 5.6278 |
| 20 | 9 | 1 | 9 | 5 | 5 | 6.305 |
| 21 | 1 | 1 | 5 | 1 | 5 | 1.9066 |
| 22 | 1 | 1 | 5 | 1 | 5 | 1.9066 |
| 23 | 1 | 1 | 1 | 1 | 5 | 1.2658 |
| 24 | 1 | 1 | 5 | 1 | 1 | 1.6382 |
| 25 | 1 | 1 | 5 | 9 | 1 | 2.3262 |

| 26 | 1 | 1 | 1 | 1 | 5 | 1.9066 |
| 27 | 5 | 9 | 1 | 1 | 1 | 4.7634 |
| 28 | 1 | 5 | 1 | 1 | 9 | 2.5638 |
| 29 | 1 | 1 | 1 | 5 | 9 | 1.8782 |
| 30 | 1 | 1 | 9 | 5 | 5 | 2.8914 |

*X represents the feature.

Among them, the score of the group of basketball players (#1-10) is significantly higher than the group of active participants (#11-20) $p = 0.0193 < 0.05$, and significantly higher than the group of less active participants (#21-30) $p = 3.59 \times 10^{-8} < 0.05$. The group of active participants is also significantly higher than the group of less active participants, $p = 0.0486 < 0.05$. From the results, there are significant differences between the three groups, indicating that the feature extraction model based on the LESS has a certain reference value for distinguishing people with different sports abilities and identifying potential sports injury risk with liability.

## Results

The experimental equipment utilized in this paper was the Jetson Nano, while the experimental environment consisted of a quad-core ARM Cortex-A57MPCore CPU, 4GB memory, Ubuntu 18.04 operating system, and PyCharm Community 2019.

Error analysis of human posture estimation

Based on the OpenPose error analysis experiment conducted by Kimberley-Dale Ng et al. [14], the PCKh@0.5 (the ratio of correct estimate of key points) was found to be 88.9% when the confidence threshold was set to 0.2, resulting in a 7.1% loss of human key points data. With a confidence threshold of 0.4, the PCKh@0.5 improved to 91.3%, but there was an 18.5% loss of key points data. For confidence thresholds of 0.6 and 0.8, the PCKh@0.5 values were 90.9% and 93.5% respectively, but the corresponding key points losses were 36.7% and 76.2%. In this paper, a confidence threshold greater than 0.4 was chosen to analyze the experimental results of the ACL potential injury risk assessment algorithm.

## Analysis and verification of the assessment algorithm

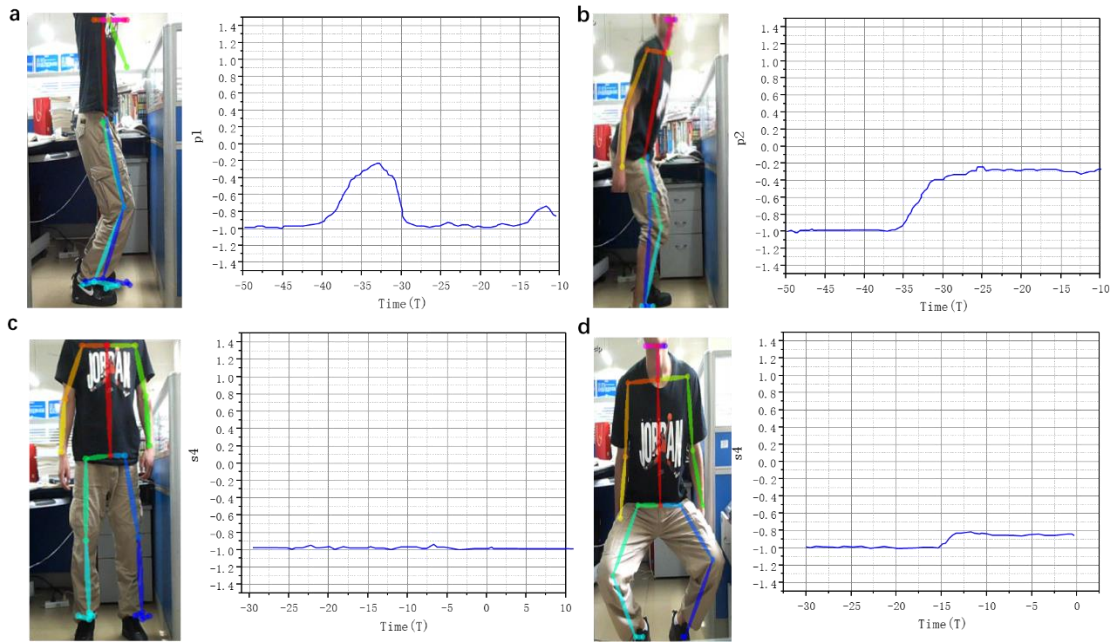

Figure 5: The first three features. (a) and (b) shows cosine of the knee flexion and hip flexion angle in the sagittal plane, respectively. (c) indicates the average and (d) represents the peak cosine value of the flexion angle of the thighs and the trunk in frontal plane.

Figure 5 provides visual representations of the experimental results during the landing stage of the platform landing test. In (a), the cosine of the knee flexion angle in the sagittal plane shows a peak value between 0 and -0.5, indicating a recorded maximum knee flexion angle close to 65 degrees, which is considered excellent. (b) illustrates the change in the cosine of the sagittal hip flexion angle, showing a peak value between 0 and -0.5, representing a recorded maximum hip flexion angle close to 70 degrees, also considered excellent. (c) displays the average flexion angle of the thighs and the trunk in the frontal plane, showing that the two are nearly parallel with a cosine value always close to -1. Finally, (d) presents the peak cosine value of the flexion angle between the two thighs and the trunk on the frontal plane, which is close to -0.75, indicating a flexion angle between 30 degrees and 60 degrees, signifying a good test result.

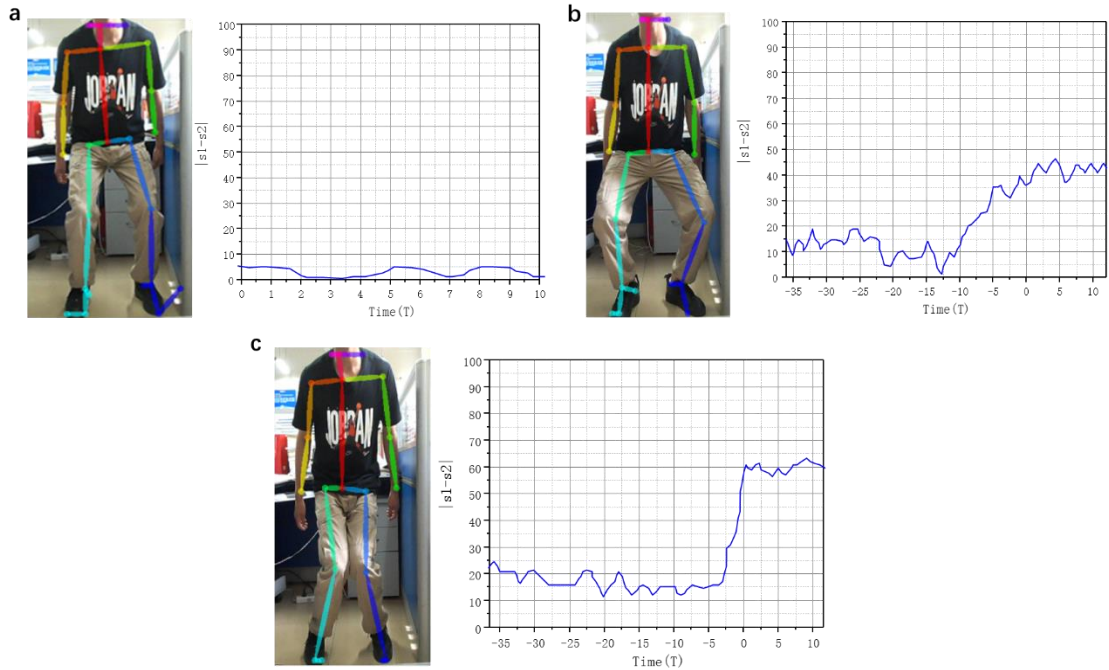

Figure 6: Different take-off postures. (a) correct take-off posture. (b) knee varus error. (c) knee valgus error.

In Figure 6, (a) represents the correct take-off posture, where the subject's take-off data are stable. In this phase, the absolute difference between the distance of the subject's knees and ankles should be less than 30°. On the other hand, (b) and (c) illustrate the errors of knee varus and knee valgus, respectively. When the knee is varus, there are fluctuations in the real-time data, and the distance between the subject's knees and ankles falls between 30 and 50. When the knee is valgus, the absolute difference between the distance of the subject's knees and ankles exceeds 50. Both knee valgus and knee varus situations are prone to ACL tears. Notably, a larger displacement in knee varus results in a more severe ACL injury.

## Conclusion

This paper introduces an ACL potential injury risk assessment method that utilizes key point detection of the human body. The study includes handling missing values in key point data, establishing a feature extraction model for ACL injury risk based on LESS, and calculating the overall score of the subject using a weighted scoring model designed with the AHP method. Through error analysis of the OpenPose algorithm and feature value comparison among three subject types, the proposed ACL injury risk assessment method based on LESS demonstrates its effectiveness in identifying potential anterior cruciate ligament injury risks in individuals.